# Decadal Temperature Prediction via Chaotic Behavior Tracking

Jinfu Ren[1], Yang Liu[1], Jiming Liu[1*]

[1]Department of Computer Science, Hong Kong Baptist University, Hong Kong SAR, China
*Correspondence to: J. Liu (jiming@comp.hkbu.edu.hk)

**Abstract**
Decadal temperature prediction provides crucial information for quantifying the expected effects of future climate changes and thus informs strategic planning and decision-making in various domains. However, such long-term predictions are extremely challenging, due to the chaotic nature of temperature variations. Moreover, the usefulness of existing simulation-based and machine learning-based methods for this task is limited because initial simulation or prediction errors increase exponentially over time. To address this challenging task, we devise a novel prediction method involving an information tracking mechanism that aims to track and adapt to changes in temperature dynamics during the prediction phase by providing probabilistic feedback on the prediction error of the next step based on the current prediction. We integrate this information tracking mechanism, which can be considered as a model calibrator, into the objective function of our method to obtain the corrections needed to avoid error accumulation. Our results show the ability of our method to accurately predict global land-surface temperatures over a decadal range. Furthermore, we demonstrate that our results are meaningful in a real-world context: the temperatures predicted using our method are consistent with and can be used to explain the well-known teleconnections within and between different continents.

**Introduction**
Accurately temperature predictions over decades[1,2,3] would enable determination of the potential effects of climate changes on areas such as food security, health system capacity, and biodiversity, thereby informing planning for agricultural production[4], responses to extreme weather events[5], and strategies for species protection[6], respectively. For example, plans to develop new crop varieties or purchase new irrigation equipment[7] could be made several years in advance to protect the yields and quality of crops, which are vulnerable to abnormal temperature fluctuations[8,9] and thereby avoid billions of dollars in losses[10,11,12]. Moreover, an understanding of temperature variability on a fine scale, e.g., weekly, is desired in long-term forecasting, as such variability greatly influences decision-making on matters such as the best periods for crop planting and harvest[13,14]. However, the determination of such fine-scale temperature dynamics across decadal periods remains a challenging research question.

Simulation-based methods (i.e., differential equation-based models) simulate climate systems using a set of differential equations and have been widely adopted for decadal temperature prediction[15,16,17,18]. Numerical weather prediction (NWP) models, the most classical form of simulation-based methods, focus on regional scales and employ some assumptions (e.g., the hydrostatic assumption[19]) to reduce the computational complexity[20,21]. Research on global circulation modeling (GCM) has increased since the 1970s[22]. In a pioneering study of GCM, Manabe et al. extended the climate modeling from the regional to the global scale by integrating the ocean system into NWP models[23]. Subsequently, GCM has been coupled with other climate-related systems, including land surface, hydrology, and vegetation systems, to yield more comprehensive models[20].

One of the primary challenges encountered by simulation-based methods is the uncertainty of temperature variations due to the chaotic nature of climate systems, as an infinitesimal difference in the initialization of equations results in an exponential increase in the error after a finite number of time steps[24,25,26]. To address this challenge, Palmer et al. developed ensemble methods[27,28] that combine predictions from multiple individual simulation-based experiments conducted under slightly different initial conditions to reduce the uncertainty in temperature variations. The Coupled Model Intercomparison Project (CMIP)[29], organized by the World Climate Research Programme (WCRP), integrates various climate models with ensemble methods to yield state-of-the-art simulations[30]. However, ensemble methods require frequent re-calibration using recent observations, which are not available for long-term forecasting tasks. Therefore, ensemble methods yield less reliable decadal temperature predictions[31].

Recently, researchers have considered machine learning (ML)-based methods as an alternative solution to the problem of temperature prediction. Such methods do not predefine models but rather exploit the power of learning models from data, which enables them to achieve competitive levels of performance in short-term prediction, compared with state-of-the-art ensemble methods[32,33]. However, even modern ML methods have a limited ability to generate models capable of making long-term predictions [34,35,36,37], as such methods require the latest ground-truth as input to avoid the exponential accumulation of prediction errors.

In this paper, we devise a novel method to address the challenge of error accumulation in decadal temperature prediction, which is faced by both simulation-based and ML-based methods. Fig. 1 illustrates the idea of the proposed method. The method comprises two components: a dependency learning component and an information tracking component. The dependency learning component, as the basis of the method, learns a predictive model from historical data. The information tracking component is more important, as it is designed to reduce uncertainty in the prediction. Specifically, this component tracks changes in temperature variation during the prediction phase by providing feedback that consists of a probabilistic estimation of the prediction error in the next step based on the current prediction. We use the first-order difference, which is generally used to quantify the Lyapunov exponent in chaotic systems, as the label to supervise the information tracking process. The feedback is integrated into the model as part of the objective function, enabling the model to make necessary corrections to its predictions. The information tracking component can thus be considered as a calibrator that prevents amplification of the initial error. From an ML perspective, the information tracking component acts as a regularizer that allows the model to adapt to the complex dynamics of a temperature time series.

To validate the effectiveness of our novel method, we test its performance on a global decadal temperature prediction task. The results of this test demonstrate that our method generates a model that can accurately predict temperatures on a weekly basis. We further visualize the results of our prediction for the 8 countries identified by the Climate Risk Index (CRI)[38], the Sustainable Development Index (SDI)[39], and the World Bank's Turn Down the Heat series[40] as being of the most concern regarding possible future temperature increases. Moreover, we validate our findings using well-known teleconnections, specifically the temperature variability relations within and between Antarctica, Asia, North America, and Europe[41,42], and show that our predictions are meaningful in a real-world context. To guarantee the ability of our method to predict a chaotic time series, we prove that it does not cause exponential error accumulation and can maintain error at a constant level during the prediction phase via the information tracking component. Please refer to the supplementary materials for details of this proof.

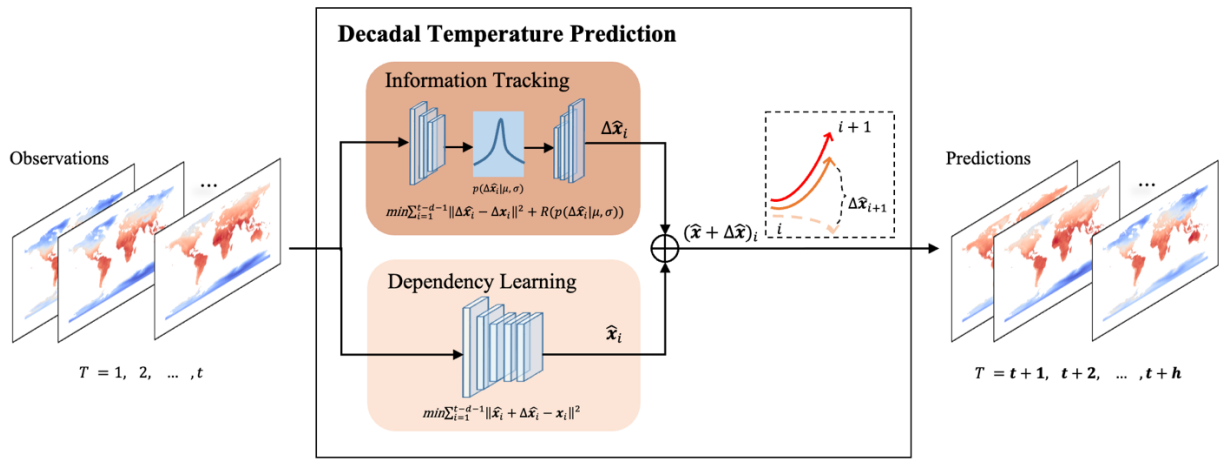

*Fig.1 **The schematic illustration of our method**. The method takes past temperature records as training data to learn the model and uses the learned model to predict the future temperature over a decadal range. It comprises two components: a dependency learning component and an information tracking component. In each step of the prediction, dependency learning component will learn the spatiotemporal dependency of the temperature time series data for target variable prediction; while the information tracking component aims to prevent the accumulation of prediction errors generated from the dependency learning component by providing a feedback of the estimated prediction error, quantified by the first-order difference of the target variable. With the compensation in each step, as shown in the box with dashed borders, the proposed method is designed to avoid exponential increase of error and provide accurate decadal temperature prediction.*

**Global Decadal Temperature Prediction**

To validate the effectiveness of the proposed method, we test its performance on a global decadal temperature prediction task. Specifically, we use the data of global land temperature from 1880 to 2010 with the resolution of 1 degree × 1 degree for training, and then we predict the next ten years' (i.e., 2011~2020) global weekly temperature. In the main manuscript, we demonstrate the prediction result for the last year (i.e., 2020) as it is the year farthest to the years of training data, presenting great challenges to the prediction task. In Fig. 2(a), we illustrate the predicted (right) and real (left) global temperature of the first week of January (top) and that of the first week of July (bottom), as the temperature in these two time slots shows great difference. The prediction results of other weeks/months/years are provided in the supplementary material. We can observe that our method is able to generate accurate predictions of decadal temperatures at the global scale, even if the temperature records show obvious heterogeneities in different geographical locations and time periods.

To demonstrate the prediction result on a finer spatiotemporal scale, we select the most concerned countries according to three important temperature-related criteria: Climate Risk Index (CRI), Sustainable Development Index (SDI), and the World Bank climate report—Turn Down the Heat series. For each criterion, we select the top 10 concerned countries, i.e., for the former two indices, we select the top 10 countries with the highest CRI and those with the lowest SDI, respectively. The former gives us the result of Puerto Rico, Myanmar, Haiti, Philippines, Mozambique, Bahamas, Bangladesh, Pakistan, Thailand, and Nepal, the latter gives us the result of India, Myanmar, Viet Nam, Bangladesh, Pakistan, Papua New Guinea, Liberia, Haiti, Turkey, and Sudan. The climate report gives results of what the average warming will be in different regions if the world warms by 4 degrees by the end of the century. We select the countries that will warm over 7.5 degrees; they are Algeria, Saudi Arabia, Iraq, Russia, China, Mongolia, Ukraine, Bulgaria, Georgia, and Romania. Four countries are repeated among 30 countries; thus 26 countries are selected finally. For each country, we select the geographical location that is the closest to the capital for demonstration. The selected countries comprehensively cover most of the continents (except for Antarctica), and the selected locations have various patterns of temperature variations.

We show the results of 8 countries in Fig. 2(b), which demonstrate certain irregularities/non-smoothness and present challenges for prediction. The results of the remaining 18 countries are presented in the supplementary material. For each country, the yellow series represent the ground truth with a duration of ten years and the green ones are our predictions; the contour of the country is given on the left with a black point indicating the geographical location that has been used for demonstration. Countries located in the North Hemisphere are in the season that is opposite to the countries located in the South Hemisphere (e.g., Mozambique versus others), so their temperature patterns have opposite trends. Moreover, we can observe that for countries located in the Tropics, e.g., Haiti, and Liberia, more noise is included, i.e., the sawtooth shape is obvious. For example, the temperature pattern of Haiti has the most obvious sawtooth shape among the selected countries in North America. The same situation happens in Liberia also, which is located in Africa, and also the other two countries Mozambique and Sudan, but not so obvious. Interestingly, the temperature trend of Sudan and Bangladesh both have a peak that has a longer range than others, due to their long hot season (Mar to September)[43,44], but the former shows the pattern of two peaks in one phase (i.e., one-year duration), and is distinct from others. In addition to that, Liberia also has a completely different temperature trend, the peak is sharper which implies the temperature change may be faster. This is consistent with the reality: only two (rather than four) seasons are in Liberia, i.e., dry and rainy seasons[45]. The great difference has been demonstrated by the countries in Southeast Asia, whose temperature patterns contain a sawtooth shape and various trends. For instance, Myanmar, Thailand, and the Philippines have totally different patterns. For Thailand, the temperature trends have a steeper downtrend. This is because Thailand has three seasons, after a hot season, there will be a rainy season and a dry season, where the temperature will drop obviously[46]. For Myanmar, except for the sawtooth shape, its temperature trend is also different from all others, which has two peaks in one phase, due to the relatively high temperature in the hot and wet seasons[47].

We zoom in to the prediction details over a two-year range in Myanmar and Sudan (the fifth row of Fig. 2(b)). Although the temporal patterns of the temperature series during such a two-year period show great irregularities and complexity, our method still can well capture the trend and make accurate predictions. Moreover, we further use a half-a-year period to demonstrate the working mechanism of information tracking in the last two rows of Fig. 2(b). As we mentioned previously, the actual first-order difference of the temperature time series (the yellow sequence in the penultimate row) is used as the label to guide the information tracking. With such a supervision, the information tracking component is able to tell the dependency learning component what will be the most likely direction of the change of the next step in the prediction phase (the blue sequence in the last row), thus together making accurate predictions on chaotic time series.

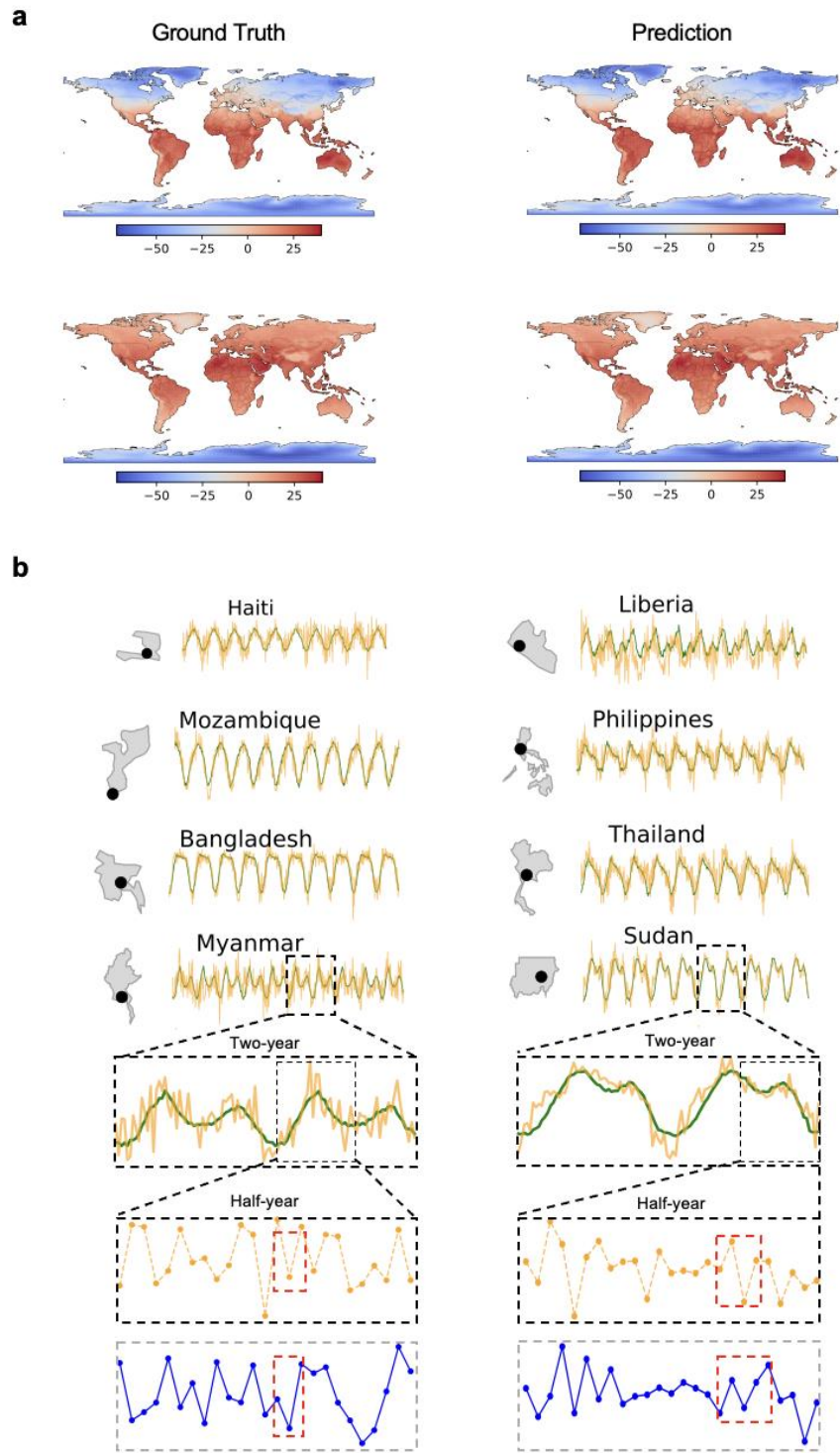

*Fig. 2 **Global decadal temperature prediction**. (a) First row: the real (left) and predicted (right) temperature in the 1st week of January 2020. Second row: the real (left) and predicted (right) temperature in the 1st week of July 2020. The red color indicates higher temperatures and the blue color indicates lower temperatures. (b) <u>The first four rows</u>: Actual (yellow) and predicted (green) temperature at the geographical locations (closest to the countries' capital) of eight selected countries that have various temperature patterns over a decadal range (2011-2020). <u>The fifth row</u>: An enlarged snippet of the prediction details over a two-year range (2016-2017) in Myanmar and Sudan. <u>The last two rows</u>: The effect of the information tracking mechanism over a half-year range. The penultimate row and the last row give the actual first-order difference of the temperature sequence and the estimated changes generated by the information tracking component, respectively.*

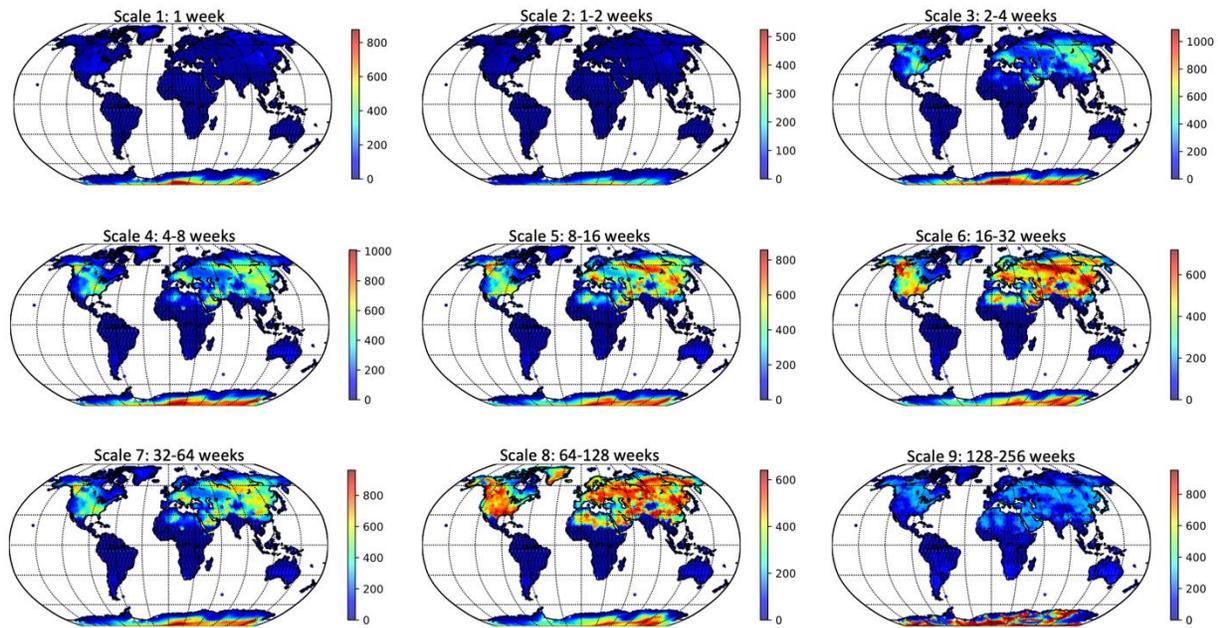

*Fig. 3 **Heat map of the number of links of each location (to other locations) at all temporal scales (from weeks to years)**. The blue and red colors represent the two extreme situations of the number. The redder, the more links the location has; the bluer, the fewer links the location has. The scales cover from week to decades, to show the change of the link density across scales.*

**Discovery of Global Teleconnections via Predicted Temperatures**
In this section, we demonstrate that our results are meaningful in a real-world context: the temperatures predicted using our method are consistent with and can be used to explain the teleconnections[49,50] in global climate, within and between different continents.

We construct the global temperature similarity matrix using the predicted results with the method developed by Agarwal et al[42]. Fig. 3 illustrates the heat map of the number of links of each location (to other locations) at all temporal scales (from weeks to years). From scale 3 (2~4 weeks), links start appearing in Asia, Europe, and North America. This becomes more obvious in the next scales (from 4~8 weeks to 64~128 weeks); for scale 9 (128~256 weeks), the links become weak, which means the teleconnection that affects the temperature between continents may not last for decades.

To further explore the connections of temperature series at different locations, we visualize the connections of one continent with others in Fig. 4 (i.e., Asia and North America), in which we uniformly select the locations to cover the whole area of the continent that we investigated. We select scale 4 for Asia and scale 8 for North America for demonstration in the main text, and put all the other scales in the supplementary material. We can observe that Asia and North America connect with each other and also both connect with the locations in the Arctic Circle, which is consistent with the fact that on the seasonal and interannual scales, for east Asia, the Arctic Oscillation, and East Atlantic/West Russia (EA/WR) could be the dominant teleconnections[51,52,53]. For example, the Arctic Oscillation depicts the dominant seesaw between the mid-latitudes of the North Hemisphere and Arctic atmospheric mass (i.e., air pressure)[54,55,56], which could modulate the movement of cold air from the Arctic (high latitudes) to East Asia that impacts the temperature change in East Asia. This is clearly shown in the connection map in Fig. 4(a), that Asia does have many connections with the locations in Arctic Circle, which demonstrates that their temperature series are highly correlated at the monthly scale (scale 4: 4~8 weeks). Similarly, for North America, Arctic Oscillation, and Asian–Bering–North American (ABNA) could be the dominant teleconnections[57,58]. For example, in ABNA, a zonally elongated wavetrain would flow into North America from North Asia through the Bering Sea, and it is a year-round phenomenon. Thus the temperature variations in North Asia and those in North America are highly correlated at the yearly scale (scale 8: 64~128 weeks) as shown in Fig. 4(b).

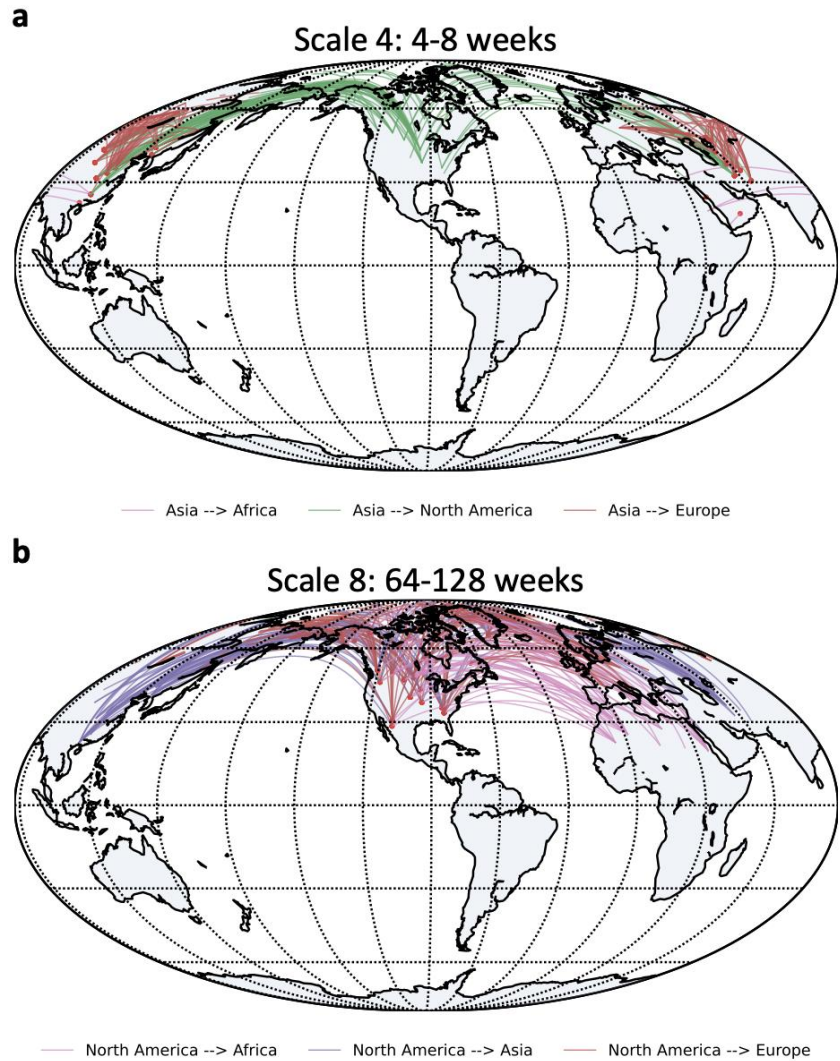

*Fig. 4 **Connection map of selected regions in the selected scales**. (a) The connections of Asia to other continents. (b) The connections of North America to other continents. Different colors represent the teleconnections between different continents.*

**Discussion**

In this work, we aim to tackle the decadal temperature prediction problem, which is challenging due to the chaotic behaviors existing in temperature series. We propose a method with the dependency learning component and information tracking component, so as to capture the complex spatiotemporal dependency and more importantly, to track the changes of temperature variations to avoid error accumulation. Results on global decadal temperature prediction demonstrate the accuracy of our method. The consistency between the predicted temperature and the real-world teleconnections further validates the effectiveness of our method in the real-world context.

The information tracking component is supervised by the first-order difference, which is, by definition, used for the calculation of the Lyapunov exponent, to describe the separation rate of two initial infinitesimally states, in other words, to characterize the stability of a dynamical system. Metric entropy is another important measurement to characterize the stability of a dynamical system and could be bounded by the sum of the positive Lyapunov exponents of the system. Thus, the first-order difference over time reflects the change in entropy to some extent if we view the temperature as the system output. From this perspective, to achieve long-term prediction, we are capturing and predicting the change in entropy at each step, which is intrinsically a characterization of how the information of the system change. That is also the reason why we named it as *information tracking*.

As mentioned before, the long-term temperature is vital in the forecast research related to sustainability. But to solve the sustainability-related issues, a broader viewpoint is necessary. For example, more climate factors could be included such as the atmosphere, topography of different areas, vegetation cover over the land, etc. As those different factors are all included in the earth system, the entropy change of the whole system may be related to each of them, i.e., the changes of different factors are coupled and interacted with each other. Thus, the design of information tracking can be extended to other scenarios when different factors are involved, by tracking their entropy change using a data-driven method when the physical knowledge is insufficient.

## Methods
### Training

The data of temperature records of the global land are from 1880 to 2020 with a resolution of 1×1 degree, and we use the data of 1880~2010 for training. For all geographical locations, we denote the data vector at the time step $i$ as $x_i = [x_i^1, x_i^2, \cdots, x_i^N] \in \mathbb{R}^{1 \times N}$, where $N$ is the number of locations and also the dimension of the vector which contains all locations' temperature records at time step $i$. Our target is to predict the next step's temperature $x_{i+1}$ by giving the previous $d$ steps temperature records. Moreover, we denote the first-order difference of the temperature series as $\Delta x_{i+1} = x_{i+1} - x_i$. We concatenate those $d$ vectors to form the $i$-th training sample, denoted as $X_i = [x_{i-d}, x_{i-d-1}, \cdots, x_i] \in \mathbb{R}^{1 \times (d*N)}$. Then we denote the output of the dependency learning component, information tracking component, and the whole model as $\hat{x}_{i+1} = DLC(X_i)$, $\Delta x_{i+1} = ITC(X_i)$, and $O_{i+1} = \hat{x}_{i+1} + \Delta x_{i+1}$, respectively. To be supervised by the ground truth of the temperature series, the output of the information tracking component is also added to the output of the dependency learning component. For the information tracking component, in order to provide feedback in a probabilistic way, the reparameterization technique is adopted. Specifically, the encoder in the component will encode the temperature vector as the hidden representation denoted as $r_i = encoder(X_i)$, then let $z$ denotes a random variable, whose approximate posterior is denoted as $q(z|\cdot)$, and follows a normal distribution $N(\mu, \sigma)$ and is parameterized by $r_i$. Here, the mean $\mu$ is parameterized by a neural network $NN_\mu$, i.e., $\mu = NN_\mu(r_i)$, and the variance $\sigma$ is parameterized by a neural network $NN_\sigma$, i.e., $\sigma = NN_\sigma(r_i)$. Each time $z$ will be sampled from the parameterized distribution and then used as input of the decoder to generate the feedback value $\Delta x_{i+1}$. The overall objective function of the model is given as follows:

$$min \sum_{i=1}^{t-d-1} (\|\hat{x}_i + \Delta \hat{x}_i - x_i\|^2 + \|\Delta \hat{x}_i - \Delta x_i\|^2) + D_{KL}(N(\mu, \sigma)\|N(\mathbf{0}, \mathbf{I})),$$

where $D_{KL}(\cdot)$ is the Kullback-Leibler divergence.

### Prediction

In the prediction phase, none of the ground truth will be involved, which means the model will generate the output based on its predicted values. For the warm-up, the first $d$ step's prediction will include part of the training data. Specifically, a sliding window with $d$ dimension will choose the last $d$ dimension data as the input of the model and generate the prediction. Then the predicted value will be concatenated with the data, and the sliding window will slide one step to choose the next step's input, i.e., a $d$ dimensional vector that contains the previous prediction. The aforementioned process will be repeated to generate the prediction and after $d$ steps, there will be no training data included at all. For each step, the prediction will generate as follows:

$$O_T = \hat{x}_T + \Delta x_T,$$

where $T$ is any time step that is not included in the training phase, $O_T$ is the model prediction of this step, which will be used as input for the next step's prediction. In this prediction, $\Delta x_T$ is the feedback value generated by the information tracking component by using the previous prediction for the dependency learning component output, which is also generated based on the same data. The summation of these two terms will be the final prediction. Different from existing prediction methods, the designed feedback mechanism in the proposed method will provide the positive feedback for the model to track the information change of the ground truth during prediction phase. We conduct one-step-ahead predictions for duration of ten years.

### Evaluation Metrics
We use three evaluation metrics: the mean absolute error (MAE), the root mean square error (RMSE), and the dynamic time wrapping (DTW) in this work. Specifically, the former two are used to evaluate the prediction error, i.e.,

$$\text{MAE} = \frac{\|O_{period} - X_{period}\|}{Nh_{period}}, \text{ and } \text{RMSE} = \frac{1}{N}\sum_{i=1}^{N}\sqrt{\frac{\|\bar{x}^i_{period} - o^i_{period}\|^2}{h_{period}}},$$

where $X_{period}$, $\bar{X}_{period} \in \mathbb{R}^{N \times d_{period}}$ are the ground truth and prediction respectively. $h_{period}$ is the dimension of the predicted timeline, $x^i_{period}$, $o^i_{period} \in \mathbb{R}^{h_{period}}$ are the prediction and ground truth of the $i$-th location respectively. In addition to the prediction error, we also aim to evaluate the results in terms of the temporal trends of the time series. For example, the difference in the phase may have the same error as a straight line does. Therefore, another metric that could describe the shape similarity would be complementary to a fair comparison. We choose dynamic time wrapping as it is widely used for time series similarity measurement.

**Availability of Data and Materials**
All codes and data will be made publicly available by the authors.

**Reference**
1. The Decadal Climate Prediction Project Overview https://www.wcrp-climate.org/dcp-overview (2017).
2. Meehl, G.A. et al. Decadal prediction: can it be skillful? *Bull. Am. Meteorol. Soc.* **90** (10), 1467-1486 (2009).
3. Meehl, G.A. et al. Initialized Earth System prediction from subseasonal to decadal timescales. *Nat. Rev. Earth Environ.* **2**(5), 340-357 (2021).
4. Solaraju-Murali, B. et al. How decadal predictions entered the climate services arena: an example from the agriculture sector, *Clim. Serv.*, **27**, 100303 (2022).
5. World Health Organization. Regional Office for Europe & European Commission. *Improving public health responses to extreme weather/heat-waves : EuroHEAT : technical summary*. (2009).
6. International Union for Conservation of Nature. *Adapting to climate change: Guidance for protected area managers and planners*. (2016).
7. Decadal predictions for agriculture overview. https://climate.copernicus.eu/decadal-predictions-agriculture (2019).
8. Lesk, C., Rowhani, P. & Ramankutty, N. Influence of extreme weather disasters on global crop production. *Nature*, **529**, 84–87 (2016).
9. Kistner, E., et al. *Vulnerability of specialty crops to short-term climatic variability and adaptation strategies in the Midwestern USA*. *Clim. Change*. **146**, 145–158 (2014).
10. Zhao, C. et al. Temperature increase reduces global yields of major crops in four independent estimates. *Proc. Natl. Acad. Sci.* **35**, 9326-9331 (2017).
11. Food and Agriculture Organization of the United Nations. *Climate Change and Food Security: Risks and Responses*. (2015).
12. Diffenbaugh, N.S., Davenport, F.V. & Burke, M. Historical warming has increased U.S. crop insurance losses, *Environ. Res. Lett.* **16** (8), 084025 (2021).
13. United States Department of Commerce and United States Department of Agriculture, *Weekly Weather and Crop Bulletin*, **110** (5), (2023).
14. United States Department of Agriculture, *Usual Planting and Harvesting Dates for U.S. Field Crops*. (2010).
15. Taylor, K.E., Stouffer, R.J. & Meehl, G.A. An overview of CMIP5 and the experiment design. *Bull. Am. Meteorol. Soc.* **93** (4), 485-498 (2012).
16. Eyring, V. et al. Overview of the Coupled Model Intercomparison Project Phase 6 (CMIP6) experimental design and organization. *Geosci. Model Dev.* **9** (5), 1937-1958 (2016).
17. Bauer, P., Thorpe, A. and Brunet, G. The quiet revolution of numerical weather prediction. *Nature*, **525** (7567), 47-55 (2015).
18. Nemytskii, V.V. et al. *Qualitative theory of differential equations*. (Princeton University Press 2015).
19. Hydrostatic Assumption. https://glossary.ametsoc.org/wiki/Hydrostatic_assumption
20. Edwards, P.N. History of climate modeling. *Wiley Interdiscip. Rev. Clim. Change*, **2** (1), 128-139 (2011).
21. Manabe, S. & Strickler, R.F. Thermal equilibrium of the atmosphere with a convective adjustment. *J. Atmos. Sci.*, **21** (4), 361-385 (1964).
22. McGuffie, K. & Henderson-Sellers, A. Forty years of numerical climate modelling. *Int J Climatol*, **21**(9), 1067-1109 (2001).
23. Manabe, S. & Bryan, K. Climate calculations with a combined ocean-atmosphere model. *J. atmos. Sci*, **26** (4), 786-789 (1969).
24. Shukla, J. Predictability in the Midst of Chaos: A Scientific Basis for Climate Forecasting, Science, **282** (5389), 728-731 (1998).

25. Slingo, J. & Palmer, T. Uncertainty in weather and climate prediction. *Philos. Trans. Royal Soc.* **369** (1956), 4751-4767 (2011).
26. Palmer, T.N., Stochastic weather and climate models. Nat. Rev. Phys. **1**, 463–471 (2019).
27. Palmer, T.N. Predicting uncertainty in forecasts of weather and climate. *Rep. Prog. Phys.* **63**(2), 71 (2000).
28. Leutbecher, M. & Palmer, T.N. Ensemble forecasting. *J. Comput. Phys.* **227** (7), 3515-3539 (2008).
29. Meehl, G.A., Boer, G.J., Covey, C., Latif, M. & Stouffer, R.J. The coupled model intercomparison project (CMIP). *Bull. Am. Meteorol. Soc.* **81**(2), 313-318 (2000).
30. Eyring, V. et al. Overview of the Coupled Model Intercomparison Project Phase 6 (CMIP6) experimental design and organization, *Geosci. Model Dev.* **9**, 1937-1958 (2016).
31. Palmer, T. The ECMWF ensemble prediction system: Looking back (more than) 25 years and projecting forward 25 years, *Q. J. R. Meteorol. Soc.*, **145**, 12-24 (2019).
32. David R. et al. Tackling Climate Change with Machine Learning. *ACM Comput. Surv.* **55** (2), 1-96 (2022).
33. Chantry, M., Christensen, H., Dueben, P. & Palmer, T. Opportunities and challenges for machine learning in weather and climate modelling: hard, medium and soft AI. *Philos. Trans. Royal Soc.* **379** (2194), 20200083 (2021).
34. Yu, R., Zheng, S., Anandkumar, A., & Yue, Y. Long-term forecasting using higher order tensor RNNs. arXiv preprint arXiv:1711.00073. (2017).
35. Pathak, J., Hunt, B., Girvan, M., Lu, Z., & Ott, E. Model-free prediction of large spatiotemporally chaotic systems from data: A reservoir computing approach. *Phys. Rev. Lett.* **120** (2), 024102 (2018).
36. Jia, X. et al. Physics guided RNNs for modeling dynamical systems: A case study in simulating lake temperature profiles. *In Proceedings of the 19th ICDM*. 558-566 (2019).
37. Saber, A., James, D. E., & Hayes, D. F. Long-term forecast of water temperature and dissolved oxygen profiles in deep lakes using artificial neural networks conjugated with wavelet transform. *Limnol. Oceanogr.* **65** (6), 1297-1317 (2020).
38. The global climate risk index 2021. (Eckstein, D., Künzel, V. & Schäfer, L., Bonn: Germanwatch , 2021).
39. Sustainable Development Index https://www.sustainabledevelopmentindex.org/ (2019).
40. Turn down the heat: Climate extremes, regional impacts, and the case for resilience. (World Bank, 2013).
41. Boers, N. et al. Complex networks reveal global pattern of extreme-rainfall teleconnections. *Nature*, **566**, 373–377 (2019).
42. Agarwal, A. et al. Network-based identification and characterization of teleconnections on different scales. *Sci. Rep.* **9** (1), 1-12 (2019).
43. Becker, S. & Weng, S. Seasonal patterns of deaths in Matlab, Bangladesh. *Int. J. Epidemiol.* **27** (5), 814-823 (1998).
44. Rahman, M.R. & Lateh, H. Climate change in Bangladesh: a spatio-temporal analysis and simulation of recent temperature and rainfall data using GIS and time series analysis model. *Theor. Appl. Climatol.* **128**, 27-41 (2017).
45. Dunbar, E.J. & Gupta, L. Temporal variation and source identification of carbonaceous aerosols in Monrovia, Liberia. *Sci. Afr.* **19**, 01540 (2023).
46. Arifwidodo, S.D. & Tanaka, T. The characteristics of urban heat island in Bangkok, Thailand. Procedia Soc. Behav. Sci. **195**, 423-428 (2015).
47. Horton, R., et al. Assessing climate risk in Myanmar: summary for policymakers and planners. New York, NY, USA: Center for Climate Systems Research at Columbia University, WWF-US and WWF-Myanmar, UNHabitat Myanmar.
48. Pohl, B., Fauchereau, N., Reason, C.J.C. & Rouault, M. Relationships between the Antarctic Oscillation, the Madden–Julian oscillation, and ENSO, and consequences for rainfall analysis. *J. Clim.* **23** (2), 238-254 (2010).
49. Cardil, A., Rodrigues, M., Tapia, M. et al. Climate teleconnections modulate global burned area. Nat Commun 14, 427 (2023).
50. Li, X. et al. Tropical teleconnection impacts on Antarctic climate changes. *Nat. Rev. Earth Environ,* **2** (10), 680-698 (2021).
51. Lim, Y.K. & Kim, H.D. Impact of the dominant large-scale teleconnections on winter temperature variability over East Asia. *J. Geophys. Res. Atmos.* **118** (14), 7835-7848 (2013).
52. Song, Y., Chen, H. & Yang, J. The Dominant Modes of Spring Land Surface Temperature Over Western Eurasia and Their Possible Linkages With Large-Scale Atmospheric Teleconnection Patterns. *J. Geophys. Res. Atmos.* **127** (4), 2021JD035720 (2022).
53. Ríos-Cornejo, D., Penas, Á., Álvarez-Esteban, R. & Del Río, S. Links between teleconnection patterns and mean temperature in Spain. *Theor. Appl. Climatol.* **122** (1), 1-18 (2015).
54. Thompson D. W. J. & Wallace J. M. The Arctic oscillation signature in the wintertime geopotential height and temperature fields. *Geophys. Res. Lett.*, **25** (9), 1297–1300 (1998).
55. He, S., Gao, Y., Li, F., Wang, H. & He, Y. Impact of Arctic Oscillation on the East Asian climate: A review. *Earth Sci. Rev.* **164**, 48-62 (2017).
56. Song, W. & Ye, X. Possible Associations between the Number of Cold Days over East Asia and Arctic Oscillation and Arctic Warming. *Atmosphere*, **12** (7), 842 (2021).

57. Yu, B., Lin, H., Wu, Z.W. & Merryfield, W.J. The Asian–Bering–North American teleconnection: seasonality, maintenance, and climate impact on North America. *Clim. Dyn.* **50** (5), 2023-2038 (2018).
58. Yu, B., Lin, H. & Soulard, N. A comparison of north american surface temperature and temperature extreme anomalies in association with various atmospheric teleconnection patterns. *Atmosphere*, **10** (4), 172 (2019).